\documentclass[letterpaper]{article} 
\usepackage{aaai24}  
\usepackage{times}  
\usepackage{helvet}  
\usepackage{courier}  
\usepackage[hyphens]{url}  
\usepackage{graphicx} 
\usepackage{natbib}  
\usepackage{caption} 
\usepackage{algorithm}
\usepackage{algorithmic}

\usepackage{newfloat}
\usepackage{listings}
\usepackage{inconsolata}
\usepackage{amsmath}
\usepackage{amsfonts}
\usepackage{amssymb}
\usepackage{mathtools}
\usepackage{amsthm}
\usepackage{bm}
\usepackage{array}
\usepackage{textcomp}
\usepackage{url}
\usepackage{verbatim}
\usepackage{multirow}
\usepackage{multicol}
\usepackage{color}
\usepackage{makecell}
\usepackage{rotating}
\usepackage{colortbl}
\usepackage{enumitem}
\usepackage{booktabs}
\usepackage{tabularx}

\DeclareCaptionStyle{ruled}{labelfont=normalfont,labelsep=colon,strut=off} 
\floatstyle{ruled}
\newfloat{listing}{tb}{lst}{}
\floatname{listing}{Listing}
\title{ZO-AdaMU Optimizer: Adapting Perturbation by the Momentum and Uncertainty in Zeroth-order Optimization}
\author{
    Shuoran Jiang\textsuperscript{\rm 1},
    Qingcai Chen\textsuperscript{\rm 1,2}\footnote{Corresponding Authors.},
    Youcheng Pan\textsuperscript{\rm 2}$^\ast$,
    Yang Xiang\textsuperscript{\rm 2},\\
    Yukang Lin\textsuperscript{\rm 1},
    Xiangping Wu\textsuperscript{\rm 1},
    Chuanyi Liu\textsuperscript{\rm 3,2},
    Xiaobao Song\textsuperscript{\rm 3}
}
\affiliations{
    \textsuperscript{\rm 1}School of Computer Science and Technology, Harbin Institute of Technology (Shenzhen), Shenzhen, China\\
    \textsuperscript{\rm 2}Peng Cheng Laboratory, Shenzhen, China\\
    \textsuperscript{\rm 3}Institute of Data Security, Harbin Institute of Technology (Shenzhen), Shenzhen, China\\

    shuoran.chiang@gmail.com, qingcai.chen@hit.edu.cn, panyoucheng4@gmail.com, xiangy@pcl.ac.cn
}

\usepackage{bibentry}

\begin{document}

\maketitle

\begin{abstract}
Lowering the memory requirement in full-parameter training on large models has become a hot research area.
MeZO fine-tunes the large language models (LLMs) by just forward passes in a zeroth-order SGD optimizer (ZO-SGD),
demonstrating excellent performance with the same GPU memory usage as inference.
However,
the simulated perturbation stochastic approximation for gradient estimate in MeZO leads to severe oscillations and incurs a substantial time overhead.
Moreover,
without momentum regularization,
MeZO shows severe over-fitting problems.
Lastly,
the perturbation-irrelevant momentum on ZO-SGD does not improve the convergence rate.
This study proposes ZO-AdaMU to resolve the above problems by adapting the simulated perturbation with momentum in its stochastic approximation.
Unlike existing adaptive momentum methods,
we relocate momentum on simulated perturbation in stochastic gradient approximation.
Our convergence analysis and experiments prove this is a better way to improve convergence stability and rate in ZO-SGD.
Extensive experiments demonstrate that ZO-AdaMU yields better generalization for LLMs fine-tuning across various NLP tasks than MeZO and its momentum variants.
\end{abstract}

\section{Introduction}

Large-scale models demonstrate remarkable abilities such as emergence and grokking \cite{wei2022emergent},
especially the large language models (LLMs) show excellent in-context learning (ICL) abilities and revolutionize the dominant methodology in various natural language processing (NLP) tasks.
However,
full-parameter fine-tuning with billions of parameters raises the bar for most NLP researches \cite{lv2023full, li2023lmeye}.
\citet{malladi2023fine} experimentally proved that back-propagation optimization necessitates approximately 12 times of memory cost of forward inference.
full-parameter fine-tuning a 6.7 billion (B) OPT \cite{zhang2022opt} with Adam \cite{KingBa15} requires at least $3\times$ A100 GPUs (240GB memory).

Therefore, 
the memory-efficient fine-tuning method has become an important research topic in the large-scale model era.
Some approaches have been proposed,
such as ICL does not need optimization \cite{sun2023does} and quickly adapts LLMs to specific use cases through demonstrations before test examples.
However,
a few demonstrations can only cover some possible case types due to the limited maximum sequence length for LLMs inputs.
The parameter-efficient fine-tuning methods (PEFT) \cite{fu2023effectiveness},
e.g., LoRA \cite{hu2021lora}, ZeRA \cite{rajbhandari2020zero}, EMAT \cite{wu2022efficient} et. al.,
update only a fraction of the model parameters.
Even though these methods can tune LLMs with low memory and computation resources,
there are more practical solutions to fully use the emergent ability as full-parameters fine-tuning \cite{ding2022delta, sun2023comparative}.

The memory-efficient full-parameter fine-tuning method is a better way to exploit the deployment of LLMs on limited resources.
The low-memory optimization (LOMO) gives up gradient storage in stochastic gradient descent (SGD) \cite{shamir2013stochastic},
but computes gradients at each training step.
Zeroth-order optimization (ZO) relies solely on forward passes to estimate gradients and update parameters,
costing the same memory size as inference.
\citet{malladi2023fine} proposed the memory-efficient zeroth-order optimizer (MeZO) to fine-tune LLMs with just forward passes,
and they achieved excellent performance on various NLP tasks.
The surprising performance of MeZO is attributed to the small local effective rank of Hession parameter matrices in pre-trained deep neural networks \cite{papyan2018full, papyan2020traces, ghorbani2019investigation, yao2020pyhessian, wu2020dissecting, Sagun2017EmpiricalAO}.
MeZO computes gradients from two forward passes with random perturbations.
Therefore,
MeZO greatly reduces the latency between GPU calculations compared to LOMO.
However,
the simulated perturbation stochastic approximation for the gradient in MeZO is non-smoothness in consecutive training steps.
Without the regularization from momentum,
like in back-propagation optimization methods,
MeZO causes the over-fitting problem.
ZO-AdaMM \cite{chen2019zo} first efforts to integrate adaptive momentum methods with ZO optimization,
but the perturbation-irrelevant momentum requires longer training steps to convergence.

Motivated by these challenges,
we propose the ZO-AdaMU optimizer,
in which we first make an effort to introduce adaptive momentum and uncertainty on simulated perturbation to approximate gradients.
Specifically,
we relocate the momentum from gradient to simulated perturbation,
which aims to improve the smoothness between gradient approximation and actual parameter moving.
In addition,
the simulated perturbation stochastic approximation (SPSA) \cite{maryak2001global} includes two parts sampled from momentum-centered and zero-centered Gaussian distribution.
The momentum-centered Gaussian in SPSA includes a uncertainty and its value is set by a simulated annealing function.
With the introduction of adaptive momentum and uncertainty,
ZO-AdaMU demonstrates high stability and faster convergence speed.
Our convergence analysis and comprehensive experiments prove these for stable convergence rates and better global convergence.
As some gradients in ZO-AdaMU inevitably deviate from the prediction,
we propose a second-order momentum approximation to improve the convergence rate further.
The details of ZO-AdaMU are summarized in Algorithm \ref{alg:auzo}.

Contributions of our paper can be summarized as follows:

\begin{itemize}
    \item Motivated by the problems of oscillation in MeZO and perturbation-irrelevant momentum in ZO-AdaMM, we first make an effort to explore a way to adapt momentum and uncertainty in the zeroth-order oracle, called ZO-AdaMU.
    \item Our theoretical analysis proved that momentum with uncertainty and its relocation to simulated perturbation improved the convergence rate. Meanwhile, these also contribute to a better local optimal point than a well-tuned MeZO counterpart.
    \item Our comprehensive experiments evaluate a variety of NLP tasks and demonstrate that ZO-AdaMU shows a faster convergence rate and better generalization capability than MeZO, LOMO, and Adam fine-tuned full-parameter LLMs.
\end{itemize}

\section{Preliminaries}

Zeroth-order optimization (ZO) is a gradient-free method,
and it estimates the gradient via the simulated perturbations stochastic approximation (SPSA).
This section briefly introduces the multi-point estimate version as in MeZO.
In addition,
the momentum-based regularization methods for ZO optimization are also described.

\subsection{Zeroth-Order Optimization}

Zeroth-order (ZO) optimization has long been studied in the context of convex and strongly convex objectives.
A classical ZO gradient estimator is a simultaneous perturbation stochastic approximation (SPSA) \cite{maryak2001global},
and it can replace the gradient calculation in stochastic gradient descent (ZO-SGD) \cite{ghadimi2013stochastic}.

Consider a labelled dataset $\mathcal{D}=\left\{ (\bm{x}_i, \bm{y}_i) \right\}_{i\in \left\| \mathcal{D} \right\|}$,
a minibatch $\mathcal{B} \subset \mathcal{D}$ of size $B$,
and let $\mathcal{L}(\boldsymbol{\theta}; \mathcal{B})$ represents the loss on the minibatch for a model with parameters $\boldsymbol{\theta}\in \mathbb{R}^d$.
The ZO gradient estimate via SPSA is defined as,

\begin{equation}
    \begin{aligned}
        \widehat{\nabla } \mathcal{L} \left( \boldsymbol{\theta };\mathcal{B} \right) &=\frac{\mathcal{L} \left( \boldsymbol{\theta }+ \epsilon \bm{z}; \mathcal{B} \right) - \mathcal{L} \left( \boldsymbol{\theta }-\epsilon \bm{z};\mathcal{B} \right)}{2 \epsilon} \bm{z} \\
        & \approx \bm{z}\bm{z}^{\top} \nabla \mathcal{L} \left( \boldsymbol{\theta };\mathcal{B} \right)
    \end{aligned}
\end{equation}
where $\bm{z}\in \mathbb{R}^d$ with $\bm{z}\sim \mathcal{N}(0, \bm{I})$, $\boldsymbol{I} \in \mathbb{R}^{\| \boldsymbol{\theta} \|}$ and $\epsilon$ is the perturbation scale.

SPSA requires only two forward passes through model $\boldsymbol{\theta}$ to estimate the gradient.
SPSA estimated gradients can be used to replace the gradient computation in any back-propagation optimizer such as SGD.

\begin{equation}
    \boldsymbol{\theta}_{t+1} = \boldsymbol{\theta}_t - \eta \widehat{\nabla} \mathcal{L}(\boldsymbol{\theta}; \mathcal{B}_t)
\end{equation}
where $\mathcal{B}_t$ denotes the monibatch at step $t$ and $\widehat{\nabla}$ is the SPSA estimated gradient.

\subsection{Memory-efficient ZO-SGD (MeZO)}

MeZO \cite{malladi2023fine} is an in-place implementation of ZO-SGD,
and it further reduces memory requirements with the same memory usage as the inference.
Specifically,
MeZO keeps the random seed $s$ for all random vector $\boldsymbol{z}$ sampling to generate each perturbation $\bm{z}$ at each step.
{\footnotesize
\begin{equation}
    \begin{aligned}
        & \mathcal{B} \in \mathcal{D} , \qquad s\gets \texttt{rand}\left(  \right) \\
        & \boldsymbol{\theta }\gets \texttt{PerturbParameters}\left( \boldsymbol{\theta },\epsilon ,s \right) \\
        & \ell _+\gets \mathcal{L} \left( \boldsymbol{\theta },\mathcal{B} \right) \\
        & \boldsymbol{\theta }\gets \texttt{PerturbParameters}\left( \boldsymbol{\theta },-2\epsilon ,s \right) \\
        & \ell _-\gets \mathcal{L} \left( \boldsymbol{\theta },\mathcal{B} \right) \\
        & \boldsymbol{\theta }\gets \texttt{PerturbParameters}\left( \boldsymbol{\theta },\epsilon ,s \right) \\
        & \texttt{grad} \gets \left( \ell _+-\ell _- \right) /\left( 2\epsilon \right) \\
        & \boldsymbol{z} \sim \mathcal{N}(0,1) \quad \texttt{with seed $s$} \\
        & \boldsymbol{\theta} \leftarrow \boldsymbol{\theta} - \eta \ast \texttt{grad} \ast \boldsymbol{z}
    \end{aligned}
\end{equation}}where $s \leftarrow \texttt{rand}()$ samples a random seed keeping same in one gradient updating step,
$\eta$ is the learning rate and $\epsilon$ is the perturbation scale.

\subsection{Adaptive Momentum Method for ZO}

ZO-AdaMM \cite{chen2019zo} first make an effort to integrate adaptive momentum methods with ZO-SGD,
and it shows theoretically convergence guarantees for both convex and non-convex constrained optimization.
The method is inspired by the Adam optimization algorithm and extends its capabilities to handle scenarios where gradient information is unavailable.
Zo-AdaMM leverages both the zeroth-order moments, which capture the function's behavior, and the historical information of previous iterations to dynamically adjust the step size and momentum for optimization.

This method has the potential to advance optimization techniques in scenarios where gradient information is not available,
opening new avenues for solving complex real-world optimization problems.
However,
it suffers a slowdown factor $\mathcal{O}(\sqrt{d})$ compared with first-order Adam.

\section{Methodology}

In this section,
we propose the ZO-AdaMU optimizer by adapting perturbation with the momentum and uncertainty in the zeroth-order oracle.
Unlike the post-hoc momentum estimate in back-propagation optimization methods,
ZO-AdaMU adapts the simulated perturbation with momentum and introduces the adaptive uncertainties in stochastic approximation.
In addition,
we propose a simulated annealing mechanism on uncertainty in SPSA and smoothing parameters to balance the weight of momentum in gradient approximation.
We theoretically analyze that ZO-AdaMU has a faster convergence rate and better global convergence.

\subsection{Adapting Momentum by Momentum and Uncertainty in SPSA}

The simulated perturbation in ZO-AdaMU is computed from two parts of a momentum-centered and a zero-centered Gaussian distribution:

{\footnotesize
\begin{equation}
    \begin{aligned}
        \dot{\boldsymbol{z}}_{t+1} &\sim \mathcal{N} \left(0, \sqrt{\alpha_{t+1}} \right) \\
        \ddot{\boldsymbol{z}}_{t+1} &\sim \mathcal{N} \left( \boldsymbol{m}_t, \sqrt{1 - \alpha_{t+1}} \right) \\
        \boldsymbol{m}_{t+1} &= \beta_1 \dot{\boldsymbol{z}}_{t+1} + (1-\beta_1) \ddot{\boldsymbol{z}}_{t+1} \\
        & \text{s.t.} \quad 0 \leqslant \alpha_{t+1} \leqslant 1, \quad 0 \leqslant \beta_1 \leqslant 1
    \end{aligned}
\end{equation}}
where $\boldsymbol{m}_t$ and $\alpha_{t+1}$ denote the momentum of history perturbations and the adaptive uncertainty,
these variables also imply the mean and variance for momentum-centered Gaussian distribution.
Consistently,
$0$ and $1-\alpha_{t+1}$ represent the mean and variance for zero-centered Gaussian distribution.
$\dot{\boldsymbol{z}}_{t+1}$ and $\ddot{\boldsymbol{z}}_{t+1}$ represent the momentum with uncertainty and purely stochastic perturbation in SPSA.
The hyper-parameter $\beta^1_{t+1}$ is the smoothing parameter,
and it is also adaptive with a simulated annealing function.

In this way,
the gradient on the minibatch $\mathcal{B}_t \in \mathcal{D}$ for a given model $f(\boldsymbol{\theta})$ is estimated by,

{\footnotesize
\begin{equation}
    \widehat{\nabla } \mathcal{L} \left( \boldsymbol{\theta};\mathcal{B}_t \right) =\frac{\mathcal{L} \left( \boldsymbol{\theta } + \epsilon \boldsymbol{m}_t; \mathcal{B}_t \right) -\mathcal{L} \left( \boldsymbol{\theta }- \epsilon \boldsymbol{m}_t;\mathcal{B}_t \right)}{2\epsilon}\boldsymbol{m}_t
\end{equation}}

ZO-AdaMU also mimics the exponential moving average (EMA) on the square of the gradient \cite{zhuang2020adabelief} to regularize step size,
and proposes a substitute defined as follows,

{\footnotesize
\begin{equation}
    \begin{aligned}
        \boldsymbol{v}_{t+1} &= \beta^2_{t+1} \dot{\boldsymbol{z}}_{t+1}^2 + (1-\beta^2_{t+1}) \ddot{\boldsymbol{z}}_{t+1}^2 \\
        \boldsymbol{\theta}_{t+1} &= \boldsymbol{\theta}_t - \eta \frac{\widehat{\nabla} \mathcal{L}(\boldsymbol{\theta}, \mathcal{B}_t)}{\sqrt{\boldsymbol{v}_{t+1}^2 + \sigma}}
    \end{aligned}
\end{equation}}
where $\beta^2_{t+1}$ is an adaptive smoothing parameter and $\sigma$ is a small noise and typically set as $10^{-8}$.

\begin{figure}[htbp]
    \centering
    \includegraphics[width=6cm]{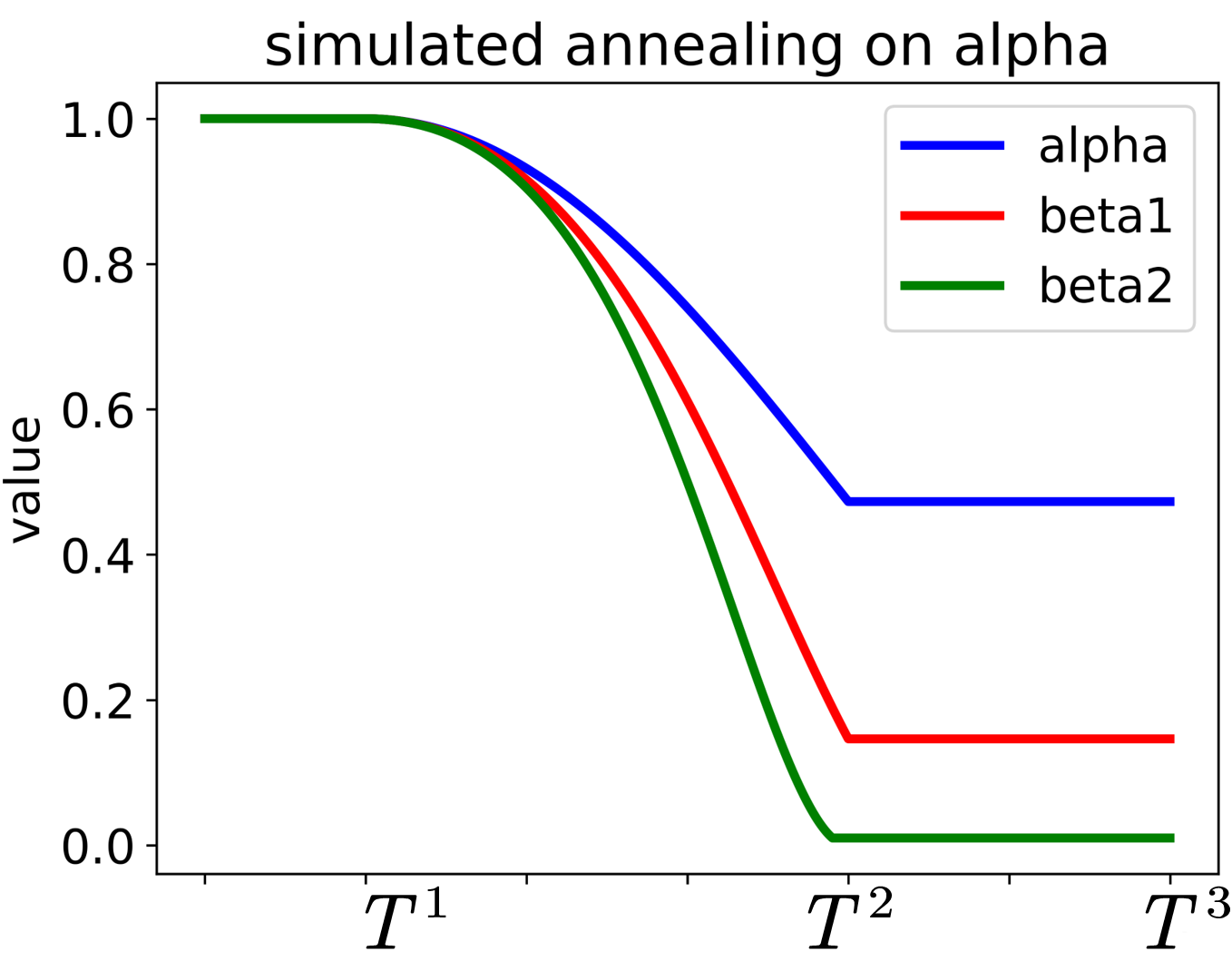}
    \caption{The simulated annealing on $\alpha$, $\beta_1$ and $\beta_3$.}
    \label{fig:simulatedannealing}
\end{figure}

The big change from traditional adaptive momentum methods is that the smoothing parameters $\beta^1_{t}$, $\beta^2_{t}$ and uncertainty $\alpha_t$ are adaptive in a simulated annealing way as follows,

{\scriptsize
\begin{equation}
    \texttt{Anneal}\left( t \right) =\begin{cases}
	1, \quad t \in [1, T^1) \\
	0.5 + 0.5 \cos \left( \pi \frac{\left( T^3-T^1 \right)}{T^3-t\varphi \frac{T^3-T^2}{T^2-T^1}} \right), t \in [T^1, T^2) \\
	0.9, \quad t \in [T^{2}, T^3) \\
    \end{cases}
    \label{eq:anneal}
\end{equation}}where $\varphi=1$ for $\alpha$, $\varphi=0.1$ for $\beta_1$ and $\varphi=1.5$ for $\beta_2$.
$t \in [1, T^1)$ is the warm-up process to estimate a major optimization orientation in SPSA without any influence of momentum.
The second $t \in [T^1, T^2)$ process accelerates the optimization by momentum-based stochastic approximation,
the uncertainty $\alpha$ on momentum gradually increases until $0.5$ as shown in Figure \ref{fig:simulatedannealing}.
The last $t \in [T^2, T^3)$ process fixes $\alpha=0.5$, $\beta^{(1)}_t=0.9$ and $\beta^{(2)}_t=0.01$ to find a global convergence.
The simulated annealing is plotted in Figure \ref{fig:simulatedannealing}.

The proposed ZO-AdaMU is summarized in Algorithm \ref{alg:auzo}.

\begin{algorithm}
\caption{ZO-AdaMU Optimizer}
\label{alg:auzo}
\begin{algorithmic}[1]
    \STATE \textbf{Input:} parameters $\boldsymbol{\theta} \in \mathbb{R}^d$, loss $\mathcal{L}: \mathbb{R}^d \rightarrow \mathbb{R}$, step budget $T^1$, $T^2$, $T^3$, perturbation scale $\epsilon$, small number $\sigma=10^{-8}$, batch size $B$, momentum uncertainty $\alpha$, learning rate $\eta$, EMA of perturbation $\boldsymbol{m}$ and $\boldsymbol{v}$ EMA on its square.
    \FOR{$t=1,\cdots,T$}
        \STATE Sample batch $\mathcal{B}_t \subset \mathcal{D}$ and random seed $s$
        \STATE $\boldsymbol{\theta} \leftarrow \texttt{Perturb}(\boldsymbol{\theta}, \boldsymbol{m}^{(t)} \epsilon,s,t)$, \quad $\ell_+ \leftarrow \mathcal{L}(\boldsymbol{\theta};\mathcal{B}_t)$
        \STATE $\boldsymbol{\theta} \leftarrow \texttt{Perturb}(\boldsymbol{\theta}, \boldsymbol{m}^{(t)}, -2\epsilon,s,t)$, \quad $\ell_- \leftarrow \mathcal{L}(\boldsymbol{\theta};\mathcal{B}_t)$
        \STATE $\boldsymbol{\theta} \leftarrow \texttt{Perturb}(\boldsymbol{\theta}, \boldsymbol{m}^{(t)}, \epsilon,s,t)$, \quad $g_t \leftarrow (\ell_+ - \ell_-)/(2\epsilon)$
        \STATE Reset random seed $s$
        \FOR{$\theta_i \in \boldsymbol{\theta}$}
            \STATE $\alpha, \beta^{(t)}_1, \beta^{(t)}_2 = \texttt{Anneal}(t)$
            \STATE $\dot{\boldsymbol{z}} \sim \mathcal{N}(0, \sqrt{\alpha})$, \quad $\ddot{\boldsymbol{z}} \sim \mathcal{N}(m^{(t-1)}_i, \sqrt{1-\alpha})$
            \STATE $m^{(t)}_i \leftarrow \beta^{(t)}_1 \cdot \dot{\boldsymbol{z}} + (1-\beta^{(t)}_1) \cdot \ddot{z}$
            \STATE $v_i = \beta^{(t)}_2 \cdot \dot{z}^2 + (1-\beta^{(t)}_2) \cdot \ddot{z}^2$
            \STATE $\theta_i \leftarrow \theta_i - \eta_t \cdot \frac{g_t}{\sqrt{\boldsymbol{v} + \sigma}} \cdot m^{(t)}_i$
        \ENDFOR
    \ENDFOR
\end{algorithmic}
\begin{algorithmic}[1]
    \STATE \textbf{Subroutine} $\texttt{Perturb}(\boldsymbol{\theta}, \boldsymbol{m}, \epsilon, s, t)$
    \STATE \quad Reset random seed $s$
    \STATE \quad $\alpha^{(t)}, \beta^{(t)}_1 \leftarrow \texttt{Anneal}(t)$
    \STATE \quad $\dot{\boldsymbol{z}} \sim \mathcal{N}(0, \boldsymbol{I} \ast \sqrt{\alpha^{(t)}})$, \quad $\ddot{\boldsymbol{z}} \sim \mathcal{N}(\boldsymbol{m}, \boldsymbol{I} \ast \sqrt{1-\alpha^{(t)}})$
    \STATE \quad $\boldsymbol{\theta} \leftarrow \epsilon \ast (\beta^{(t)}_1 \cdot \dot{\boldsymbol{z}} + (1-\beta^{(t)}_1) \cdot \ddot{\boldsymbol{z}})$
\end{algorithmic}
\begin{algorithmic}[1]
    \STATE \textbf{Subroutine} $\texttt{Anneal}(t, T^{1}, T^{2}, T^3)$
    \STATE \quad $\alpha \leftarrow \texttt{Anneal}(t)$ \qquad \texttt{\# refer to Eq. (7)}
\end{algorithmic}
\end{algorithm}

\begin{table*}[htbp]
  \centering
    \begin{tabular}{lccccccc|cc|cc}
    \toprule
    Task  & SST-2 & RTE   & CB    & BoolQ & WSC   & WIC   & MultiRC & COPA  & ReCoRD & SQuAD & DROP \\
    Task type & \multicolumn{7}{c|}{classification} & \multicolumn{2}{c|}{multiple choice} & \multicolumn{2}{c}{generation} \\
    \midrule
    Zero-shot & 58.8  & 59.6  & 46.4  & 59.0  & 38.5  & 55.0  & 46.9  & 80.0  & 81.2  & 46.2  & 14.6 \\
    In-context & 87.0  & 62.1  & 57.1  & 66.9  & 39.4  & 50.5  & 53.1  & 87.0  & 82.5  & 75.9  & 29.6 \\
    linear probing & 93.4  & 68.6  & 67.9  & 59.3  & 63.5  & 60.2  & 63.5  & 55.0  & 27.1  & 3.7   & 11.1 \\
    \midrule
    MeZO  & 91.4  & 66.1  & 67.9  & 67.6  & 63.5  & \textbf{61.1}  & 60.1  & 88.0  & 81.7  & 84.7  & 30.9 \\
    MeZO (LoRA) & 89.6  & 67.9  & 66.1  & 73.8  & \textbf{64.4}  & 59.7  & 61.5  & 87.0  & 81.4  & 83.8  & 31.4 \\
    MeZO (prefix) & 90.7  & 70.8  & 69.6  & 73.1  & 57.7  & 59.9  & \textbf{63.7}  & 84.0  & 81.2  & 84.2  & 28.9 \\
    \midrule
    ZO-AdaMU (2$\times$)  & \textbf{92.1} & \textbf{72.9} & 67.9  & 73.0  & 61.5 & 60.7 & 63.0 & \textbf{89.0} & 83.0 & 82.4 & 32.0 \\
    ZO-AdaMU (LoRA) & 88.0 & 72.0  & 71.6  & 72.6  & 60.1 & 56.4 & 58.9 & 88.0 & 
\textbf{83.2} & 76.8 & \textbf{32.4} \\
    ZO-AdaMU (prefix) & 88.0 & 61.8 & \textbf{72.3} & \textbf{74.9} & 56.5 & 58.2 & 61.9 & 86.0 & 82.8 & \textbf{85.2} & 30.4 \\
    \midrule
    Adam (FT) (12 $\times$) & 92.0  & 70.8  & 83.9  & 77.1  & 63.5  & 70.1  & 71.1  & 79.0  & 74.1  & 84.9  & 31.3 \\
    \bottomrule
    \end{tabular}%
    \caption{Experiments on OPT-13B (with 1,000 examples).}
  \label{tab:optresult}%
\end{table*}%

\section{Convergence Analysis}

We give the theoretical analysis about why ZO-AdaMU has higher convergence rate and better global convergence.
We follow the convergence analysis in MeZO and pay more attention to why adapting perturbation with momentum and uncertainty can improve the stability of ZO-SGD.
Therefore,
this analysis highlights the positive gains on convergence rate from perturbation momentum and uncertainty.

\subsection{Stable Convergence Rate}

Classical descent lemma on SGD optimization highlights that the larger gradient covariance results in slower decrease in loss \cite{megerle2023stable}.

\textbf{Lemma 1} (Descent Lemma). \textit{Let $\mathcal{L}(\boldsymbol{\theta})$ be $\ell$-smooth \cite{wang2019differentially}.
For any unbiased gradient estimate $\boldsymbol{g}(\boldsymbol{\theta}, \mathcal{B})$,}

{\scriptsize
\begin{equation}
    \begin{aligned}
        \mathbb{E} \left[ \mathcal{L} \left( \boldsymbol{\theta }_{t-1} \right) |\boldsymbol{\theta }_t \right] -\mathcal{L} \left( \boldsymbol{\theta }_t \right) \le & -\eta \left\| \nabla \mathcal{L} \left( \boldsymbol{\theta }_t \right) \right\| ^2 \\
        & +\frac{1}{2}\eta ^2\ell \cdot \mathbb{E} \left[ \left\| \boldsymbol{g}\left( \boldsymbol{\theta },\mathcal{B} _t \right) \right\| ^2 \right]
    \end{aligned}
\end{equation}
}

The gradient norm plays important role in the descent lemma.
We derive the gradient norms for MeZO and ZO-AdaMU respectively as below.

\textbf{Lemma 2.} \textit{Let $\mathcal{B}$ be a random minibatch of size $B$, so the gradient norms of MeZO and ZO-AdaMU are}

{\scriptsize
\begin{equation}
    \left\| \widehat{\nabla }\mathcal{L} \left( \boldsymbol{\theta },\mathcal{B} \right) \right\| ^2\sim \mathcal{N} \left( \frac{d+n-1}{n}\left\| \nabla \mathcal{L} \left( \boldsymbol{\theta },\mathcal{B} \right) \right\| ,1 \cdot \epsilon ^2 \right)
\end{equation}
}
where $\epsilon$ represents the perturbation sampling scale.

Thus,
{\scriptsize
\begin{equation}
    \underset{\eta _{\textit{MeZO}}^{-}}{\underbrace{\frac{\left\| \nabla \mathcal{L} \left( \boldsymbol{\theta }_t \right) \right\| ^2}{\frac{d+n-1}{n}\left\| \nabla \mathcal{L} \left( \boldsymbol{\theta }_t \right) \right\| ^2+\epsilon}}}\leqslant \frac{\left\| \nabla \mathcal{L} \left( \boldsymbol{\theta }_t \right) \right\| ^2}{\left\| \boldsymbol{g}\left( \boldsymbol{\theta }_t,\mathcal{B} \right) \right\| ^2}\leqslant \underset{\eta _{\textit{MeZO}}^{+}}{\underbrace{\frac{\left\| \nabla \mathcal{L} \left( \boldsymbol{\theta }_t \right) \right\| ^2}{\frac{d+n-1}{n}\left\| \nabla \mathcal{L} \left( \boldsymbol{\theta }_t \right) \right\| ^2+\epsilon}}}
\end{equation}
}

In ZO-AdaMU,
the simulated perturbation includes two Gaussian distributions with variances $\alpha$ and $1-\alpha$ and smoothing parameter $\beta_1$.

{\scriptsize
\begin{equation}
    \begin{aligned}
        \left\| \widehat{\nabla }\mathcal{L} \left( \boldsymbol{\theta },\mathcal{B} \right) \right\| ^2\sim \mathcal{N} ( \frac{d+n-1}{n}\left\| \nabla \mathcal{L} \left( \boldsymbol{\theta },\mathcal{B} \right) \right\| , \\
        \left( \beta_1^2 \alpha^2 + (1-\beta_1)^2 (1-\alpha)^2 \right) \epsilon^2 )
    \end{aligned}
\end{equation}
}

So that,
{\scriptsize
\begin{equation}
    \begin{aligned}
        \eta_{\textit{AdaMU}}^{-} &\leqslant \frac{\left\| \nabla \mathcal{L} \left( \boldsymbol{\theta }_t \right) \right\| ^2}{\left\| \boldsymbol{g}\left( \boldsymbol{\theta }_t,\mathcal{B} \right) \right\| ^2}\leqslant \eta_{\textit{AdaMU}}^{+} \\
        \eta_{\textit{AdaMU}}^{-} &=\frac{\left\| \nabla \mathcal{L} \left( \boldsymbol{\theta }_t \right) \right\| ^2}{\frac{d+n-1}{n}\left\| \nabla \mathcal{L} \left( \boldsymbol{\theta }_t \right) \right\| ^2+\epsilon \sqrt{\beta_1^2 \alpha^2 + (1-\beta_1)^2 (1-\alpha)^2}} \\
        \eta_{\textit{AdaMU}}^{+} &=\frac{\left\| \nabla \mathcal{L} \left( \boldsymbol{\theta }_t \right) \right\| ^2}{\frac{d+n-1}{n}\left\| \nabla \mathcal{L} \left( \boldsymbol{\theta }_t \right) \right\| ^2-\epsilon \sqrt{\beta_1^2 \alpha^2 + (1-\beta_1)^2 (1-\alpha)^2}}
    \end{aligned}
\end{equation}
}

As $\sqrt{\beta_1^2 \alpha^2 + (1-\beta_1)^2 (1-\alpha)^2} < 1$,
we can conclude that $\eta _{\textit{MeZO}}^{-}<\eta _{\textit{AdaMU}}^{-}\leqslant \eta _{\textit{AdaMU}}^{+}<\eta _{\textit{MeZO}}^{+}$ and $\eta _{\textit{MeZO}}=\frac{n}{d+n-1}\eta _{\textit{SGD}}<\eta _{\textit{AdaMU}}$.
Therefore,
ZO-AdaMU has a faster convergence rate than MeZO optimization.

The uncertainty in simulated perturbation also decreases the local effective rank of the Hessian of the loss \cite{papyan2018full, papyan2020traces, ghorbani2019investigation, yao2020pyhessian, Sagun2017EmpiricalAO, wu2020dissecting}.

\textbf{Lemma 3.} \textit{Let {\small $G\left( \boldsymbol{\theta }_t \right)=\max _{\left( \bm{x},\bm{y} \right) \in \mathcal{B}}$ $\left\| \nabla \mathcal{L} \left( \bm{\theta }_t;\left\{ \left( \bm{x},\bm{y} \right) \right\} \right) \right\| $}, for all $\boldsymbol{\theta}_t$ such that $\left\| \boldsymbol{\theta }-\boldsymbol{\theta }_t \right\| \le \eta d G\left[ \left( \boldsymbol{\theta } \right) \right]$ there is $\nabla ^2\mathcal{L} \left( \boldsymbol{\theta } \right) \preceq \boldsymbol{H}\left( \boldsymbol{\theta }_t \right) $, therefore the maximum of effective rank of gradient is ${tr\left( \boldsymbol{H}\left( \boldsymbol{\theta }_t \right) \right)} / {\left\| \boldsymbol{H}\left( \boldsymbol{\theta }_t \right) \right\|}_{op}\approx r $.}

With the same parameter size $d$ and minibatch size $B$,
the averaged $\widehat{G}(\boldsymbol{\theta}_t)$ on gradient estimates of MeZO and ZO-AdaMU have

{\scriptsize
\begin{equation}
    \widehat{G}_{\textit{MeZO}}(\boldsymbol{\theta}_t) > \widehat{G}_{\textit{AdaMU}}(\boldsymbol{\theta}_t)
\end{equation}
}
and thus,
{\scriptsize
\begin{equation}
    \frac{tr\left( \boldsymbol{H}_{\textit{MeZO}}\left( \boldsymbol{\theta }_t \right) \right)}{\left\| \boldsymbol{H}_{\textit{MeZO}}\left( \boldsymbol{\theta }_t \right) \right\| _{op}}\leqslant \frac{tr\left( \boldsymbol{H}_{\textit{AdaMU}}\left( \boldsymbol{\theta }_t \right) \right)}{\left\| \boldsymbol{H}_{\textit{AdaMU}}\left( \boldsymbol{\theta }_t \right) \right\| _{op}}
\end{equation}
}

The above analysis proves that ZO-AdaMU has a faster speed than MeZO to decrease the loss at each step.

\subsection{Better Global Convergence}

The upper bound on expected average regret reflects whether the optimization method can converge to a local optimum \cite{shamir2017optimal, zhuang2020adabelief}.

\textbf{Lemma 4.} \textit{The convergence of SGD optimization is commonly measured by the expected average regret,}

{\scriptsize
\begin{equation}
    \mathbb{E} \left[ R\left( T \right) \right] =\sum\nolimits_{t=1}^T{\left[ f_t\left( \boldsymbol{\theta }_t \right) -f_t\left( \boldsymbol{\theta }^{\ast} \right) \right]}
\end{equation}
}where $f_t\left( \boldsymbol{\theta }^{\ast} \right)$ is the best value with optimal solution $\boldsymbol{\theta}^{\ast}$ on $t$-th step.

\begin{table*}[htbp]
  \centering
    \begin{tabular}{lcccccc}
    \toprule
    Task  & SST-2 & SST-5 & SNLI  & MNLI  & RTE   & TREC \\
    Type  & \multicolumn{2}{c}{sentiment} & \multicolumn{3}{c}{natural language inference} & \multicolumn{1}{l}{topic} \\
    \midrule
    Zero-shot & 79.0  & 35.5  & 50.2  & 48.8  & 51.4  & 32.0 \\
    \midrule
    Adam  & 91.9 ($\pm$1.8) & 47.5 ($\pm$1.9) & 77.5 ($\pm$2.6) & 70.0 ($\pm$2.3) & 66.4 ($\pm$7.2) & 85.0 ($\pm$2.5) \\
    Adam (LoRA) & 91.4 ($\pm$1.7) & 46.7 ($\pm$1.1) & 74.9 ($\pm$4.3) & 67.7 ($\pm$1.4) & 66.1 ($\pm$3.5) & 82.7 ($\pm$4.1) \\
    Adam (prefix) &  91.9 ($\pm$1.0) & 47.7 ($\pm$1.1) & 77.2 ($\pm$1.3) & 66.5 ($\pm$2.5) & 66.6 ($\pm$2.0) & 85.7 ($\pm$1.3) \\
    \midrule
    LP    &  91.3 ($\pm$0.5) & 51.7 ($\pm$0.5) & 80.9 ($\pm$1.0) & 71.5 ($\pm$1.1) & 73.1 ($\pm$1.5) & 89.4 ($\pm$0.5) \\
    MeZO  & 93.3 ($\pm$0.7) & 53.2 ($\pm$1.4) & 83.0 ($\pm$1.0) & 78.3 ($\pm$0.5) & 78.6 ($\pm$2.0) & 94.3 ($\pm$1.3) \\
    MeZO (LoRA) & 93.4 ($\pm$0.4) & 52.4 ($\pm$0.8) & 84.0 ($\pm$0.8) & 77.9 ($\pm$0.6) & 77.6 ($\pm$1.3) & 95.0 ($\pm$0.7) \\
    MeZO (prefix) & 93.3 ($\pm$0.1) & 53.6 ($\pm$0.5) & 84.8 ($\pm$1.1) & \textbf{79.8} ($\pm$1.2) & 77.2 ($\pm$0.8) & 94.4 ($\pm$0.7) \\
    MeZO-Adam & 93.3 ($\pm$0.6) & \textbf{53.9} ($\pm$0.8) & 85.3 ($\pm$0.8) & \textbf{79.6} ($\pm$0.4) & \textbf{79.2} ($\pm$1.2) & 95.1 ($\pm$0.3) \\
    \midrule
    ZO-AdaMU  & \textbf{93.8} ($\pm$0.3) & 53.7 ($\pm$0.8) & 83.3 ($\pm$0.7) & 78.5 ($\pm$1.3) & 77.9 ($\pm$ 2.0) & 93.7 ($\pm$2.5) \\
    ZO-AdaMU (LoRA) & 93.1 ($\pm$0.6) & 51.6 ($\pm$1.2) & 84.4 ($\pm$0.5) & 78.6 ($\pm$0.8) & 78.2 ($\pm$1.2) & \textbf{95.2} ($\pm$0.5) \\
    ZO-AdaMU (prefix) & 93.4 ($\pm$0.4) & 53.6 ($\pm$0.7) & \textbf{85.5} ($\pm$0.2) & 79.4 ($\pm$1.0) & 78.9 ($\pm$1.5) & 95.0 ($\pm$0.7) \\
    \bottomrule
    \end{tabular}%
    \caption{Experiments on RoBERTa-large (350M parameters) that include zero-shot learning, linear probing (LP), full-parameter fune-tuning with Adam, MeZO and ZO-AdaMU, and parameter-efficient Fine-tuning (LoRA and prefix learning) with Adam, MeZO and ZO-AdaMU respectively. All reported numbers are averaged accuracy (standard deviation) over 5 times runs.}
    \label{tab:auzoRoberta}%
\end{table*}%

Assume that the loss $\mathcal{L}(\boldsymbol{\theta})$ has bounded gradients that $\left\| \nabla \mathcal{L} _t\left( \boldsymbol{\theta } \right) \right\| ^2\le G$ and $\left\| \nabla \mathcal{L} _t\left( \boldsymbol{\theta } \right) \right\|_{\infty} \le G_{\infty}$,
and the distance between any $\theta^{\texttt{MeZO}}_t$, $\theta^{\texttt{AdaMU}}_t$ generated by MeZO and ZO-AdaMU are both bounded as $\left\| \theta ^{\texttt{MeZO}}_n-\theta ^{\texttt{MeZO}}_m \right\| _2\le D \& \left\| \theta ^{\texttt{MeZO}}_n-\theta ^{\texttt{MeZO}}_m \right\| _{\infty}\le D_{\infty}$ and $\left\| \theta ^{\texttt{AdaMU}}_n-\theta ^{\texttt{AdaMU}}_m \right\| _2\le D \& \left\| \theta ^{\texttt{AdaMU}}_n-\theta ^{\texttt{AdaMU}}_{\infty} \right\| _2\le D_{\infty}$ respectively for any $m,n \in \left\{1, \cdots, T \right\}$.
The maximum smoothing parameters $\beta_1=0.9$ and $\beta_2=0.01$ in ZO-AdaMU and $\beta_1=0$, $\beta_2=0$ in MeZO respectively.
ZO-AdaMU and MeZO achieve the following guarantees respectively, for all $T\geqslant 1$.

{\scriptsize
\begin{equation}
    \begin{aligned}
        R_{\texttt{MeZO}}\left( T \right) &\leqslant \frac{D^2}{2\alpha}\sum_{i=1}^d{\sqrt{T}}+\alpha G_{\infty}\sum_{i=1}^d{\left\| g_{1:T,i} \right\| _2}+\sum_{i=1}^d{\frac{D_{\infty}^{2}G_{\infty}}{2\alpha \lambda ^2}} \\
        R_{\texttt{AdaMU}}\left( T \right) &\leqslant \frac{D^2}{0.2\alpha}\sum_{i=1}^d{\sqrt{T v_{T,i}}}+\frac{1.9\alpha G_{\infty}}{0.099\times 7.1^2}\sum_{i=1}^d{\left\| g_{1:T,i} \right\| _2} \\
        & +\sum_{i=1}^d{\frac{D_{\infty}^{2}G_{\infty}\sqrt{1-\beta _2}}{1.8\alpha \left( 1-\lambda \right) ^2}} \\
        R_{\texttt{MeZO}}(T) &> R_{\texttt{AdaMU}}(T)
    \end{aligned}
\end{equation}
}

Above regret bound analysis shows that ZO-AdaMU has a smaller expected average regret than the on of MeZO,
which proves that ZO-AdaMU has a better global convergence than MeZO.
This is also the reason why ZO-AdaMU achieves better generalization on LMs.

\section{Experiment}

Preliminary researches \cite{brown2020language, gao2021making, schick2021exploiting} experimentally demonstrated that zeroth-order optimization only works with prompt learning on LLMs fine-tuning.
All experiments in this section use prompts to train the LLMs with just forward passes fine-tuning (MeZO \cite{malladi2023fine} and ZO-AdaMU) and back-propagation fine-tuning (Adam).

To evaluate the effectiveness of the proposed adaptive perturbation with momentum in ZO-AdaMU for LLMs fine-tuning,
we conduct the same experiments as MeZO on both masked language model (MLM) pre-trained LLMs (like RoBERTa-large 350M) \cite{liu2019roberta} and auto-regressive pre-trained LLMs (OPT-13B) \cite{zhang2022opt} in few-shot and many-shot settings with prompts.
In addition,
all optimization methods are explored on full-parameter,
LoRA and prefix fine-tuning \cite{li2021prefix}.
Finally,
we give the visualizations of ZO-AdaMU, MeZO and Adam on 6 popular test functions for optimization.

Please refer to our code \footnote{https://github.com/MathIsAll/ZO-AdaMU.git} for the details about datasets and prompt templates in Tabels \ref{tab:optresult}, \ref{tab:auzoRoberta} and \ref{tab:nondifferential}, and hyper-parameters, grid searching for best values in Eq. \ref{eq:anneal} and experimental results to evaluate stable convergence.

\subsection{Auto-Regressive Language Models}

As the auto-regressive LLMs have become the predominant base models in NLP,
like GPT-3.5, GPT-4 \cite{lin2023comparison}, LLaMA \cite{touvron2023llama} and ChatGLM \cite{du2022glm},
we conduct experiments with OPT-13B on three NLP task paradigms - sentence classification,
multiple choice and text generation.
All benchmarks are selected from SuperGLUE \cite{wang2019superglue} (includes COPA, SST-2, RTE, CB, WSC, WIC, MultiRC, ReCoRD), BoolQ \cite{clark2019boolq}, SQuAD \cite{rajpurkar2016squad} and DROP \cite{dua2019drop}.
The few-shot training, validation and test sets are randomly sampled from each dataset with numbers of $1,000$, $500$ and $1,000$ respectively.
The main results are listed in Table \ref{tab:optresult},
and which can reach the following observations and summaries.

\uppercase\expandafter{\romannumeral1}. ZO-AdaMU has obvious advantages in complex reasoning tasks.
Table \ref{tab:optresult} shows that ZO-AdaMU and its LoRA, prefix variants outperform the MeZO and Adam fine-tuned OPT on all multiple choice and text generation tasks.
Specifically,
ZO-AdaMU and its LoRA, prefix variants outperform MeZO's results with $1.0\%$, $1.0\%$ and $2.0\%$ on COPA and $1.3\%$, $1.8\%$ and $1.6\%$ on ReCoRD respectively.
Moreover,
the best F1 scores of ZO-AdaMU on SQuAD and DROP are both $1.0$ higher than the best ones of MeZO.
The advantage of ZO-AdaMU for Adam is more evident with gaps of $10.0$, $9.1$, $0.3$ and $1.1$ respectively.

\uppercase\expandafter{\romannumeral2}. ZO-AdaMU performs closest to back-propagation optimization methods across classification tasks.
Experimental results in Table \ref{tab:optresult} show that the back-propagation optimization methods have more advantages than gradient-free methods on text classification tasks.
Specifically,
ZO-AdaMU obtains 3 best results out of 7 classification benchmarks,
and beats MeZO counterparts on all tasks.

\begin{table}[htbp]
{\scriptsize
\begin{tabular}{l|p{0.25cm}p{0.25cm}|p{0.25cm}p{0.25cm}|p{0.25cm}p{0.25cm}|p{0.25cm}p{0.25cm}|p{0.25cm}p{0.25cm}}
\hline
\multirow{2}{*}{Tasks} & \multicolumn{2}{c|}{AdaMU} & \multicolumn{2}{c|}{Adam} & \multicolumn{2}{c|}{AdamW} & \multicolumn{2}{c|}{AdaMax} & \multicolumn{2}{c}{AadMM} \\ \cline{2-11} 
                       & $\delta$ & $g$ & $\delta$ & $g$ & $\delta$ & $g$ & $\delta$ & $g$ & $\delta$ & $g$ \\ \hline
SST2                   & \textbf{92.1}    & 90.6    & 90.4         & 90.0       & 88.2         & 91.3        & 87.9          & 64.8        & 90.6         & 78.3       \\
ReCoRD                 & \textbf{83.0}    & 81.2    & 73.1         & 71.3       & 83.0         & 79.1        & 77.6          & 82.0        & 80.3         & 80.4       \\
SQuAD                  & \textbf{82.4}    & 82.1    & 82.0         & 81.0       & 77.3         & 68.4        & 80.0          & 68.3        & 79.7         & 73.8       \\ \hline
\end{tabular}}
\caption{Ablation study by adapting different momentum schedules on perturbation ($\delta$) and gradient ($g$) respectively.}
\label{tab:ablation_momentum}
\end{table}

We design an ablation study (Table \ref{tab:ablation_momentum}) by adapting momentum schedules of Adam, AdamW, AdaMax, Rmsgrad on perturbation and gradients in ZO for LLMs prompt-tuning, respectively.
These results show that our momentum schedule achieves the best results.
In addition,
the momentum schedules on perturbation are generally better than the ones on gradients,
which verifies our idea that adapting momentum on the perturbation is the right way.

\subsection{Masked Language Models}

Experimental results on OPT-13B demonstrated the promising results of ZO-AdaMU on auto-regressive pre-rained LLMs.
The second experiment extends ZO-AdaMU to the RoBERTa-large,
a popular medium-size LM in the MLM family.
This experiment follows the few-shot and many-shot settings from \citet{gao2021making} and \citet{malladi2023kernel},
where $k=512$ examples are sampled from per class for many-shot fine-tuning.
The results are summarized in Table \ref{tab:auzoRoberta}.

These results evaluate that (\romannumeral1) both ZO-AdaMU and MeZO significantly outperform the zero-shot and linear probing methods,
which proves gradient-free ZOs really tune the LLMs.
(\romannumeral2) ZO-AdaMU and MeZO outperform Adam on 6 benchmarks,
which demonstrates that the ZO-SGD methods effectively alleviate the over-fitting problem when the LLMs are fine-tuned on limited training data.
(\romannumeral3) The SPSA estimated gradient in Adam shows just a $0.43$ average improvement,
while ZO-AdaMU exhibits more dramatic increases with an average of $1.25$.

The above experiments on MLM pre-trained LMs demonstrate that the concepts of adapting perturbation with momentum and uncertainty in SPSA are more suitable for ZO-SGD methods for LLMs fine-tuning.

\subsection{Non-differentiable objectives}

\begin{table}[htbp]
  \centering
  {\scriptsize
    \begin{tabular}{lccccc}
    \toprule
    Model & \multicolumn{4}{c}{RoBERTa-large (350M)} & OPT-13B \\
    Task  & \textbf{SST-2} & \textbf{SST-5} & \textbf{SNLI} & \textbf{TREC} & \textbf{SQuAD} \\
    \midrule
    Zero-shot & 79.00  & 35.50  & 50.20  & 32.00  & 46.20  \\
    MeZO & 92.70  & 48.90  & \textbf{82.70}  & 68.60  & 78.50  \\
    ZO-AdaMU & \textbf{93.40}  & \textbf{51.60}  & 82.40  & \textbf{77.50}  & \textbf{81.30}  \\
    \bottomrule
    \end{tabular}}
  \caption{ZO-AdaMU and MeZO with non-differentiable objectives. For classification tasks of SST-2, SST-5, SNLI and TREC, RoBERTa-large is optimized with full-parameter and accuracy on 500 examples; for SQuAD, OPT-13B is optimized with prefix and F1 on 1,000 examples.}
  \label{tab:nondifferential}%
\end{table}%

As our proposed ZO-AdaMU is also a gradient-free optimization method,
we also conduct an experiment to evaluate ZO-AdaMU on RoBERTa-large and OPT with accuracy or F1 as objectives.
Table \ref{tab:nondifferential} lists all results and they demonstrate that ZO-AdaMU outperforms its MeZO counterpart on 4 out of 5 non-differentiable objectives. 

\subsection{Memory usage}

\begin{table}[htbp]
  \centering
  {\scriptsize
    \begin{tabular}{lccc}
    \toprule
    \multirow{2}[2]{*}{Hardware} & \multicolumn{3}{c}{Largest OPT that can fit} \\
          & Adam  & MeZO  & ZO-AdaMU \\
    \midrule
    1$\times$A100 (80GB) & 2.7B  & 30B   & 30B \\
    2$\times$A100 (160GB) & 6.7B  & 66B   & 66B \\
    4$\times$A100 (320GB) & 13B   & 66B   & 66B \\
    8$\times$A100 (640GB) & 30B   & 175B  & 175B \\
    \midrule
    1$\times$V100 (30GB) & 2.7B   & 6.7B  & 6.7B \\
    2$\times$V100 (60GB) & 2.7B   & 13B  & 13B \\
    4$\times$V100 (120GB) & 6.7B   & 30B  & 30B \\
    8$\times$V100 (240GB) & 13B   & 66B  & 66B \\
    \bottomrule
    \end{tabular}}
    \caption{Largest OPT models that the mainstream hardwares of Nvidia A100 and V100 can tune.}
    \label{tab:memory}%
\end{table}%

\begin{figure*}[htbp]
    \centering
    \includegraphics[width=15cm]{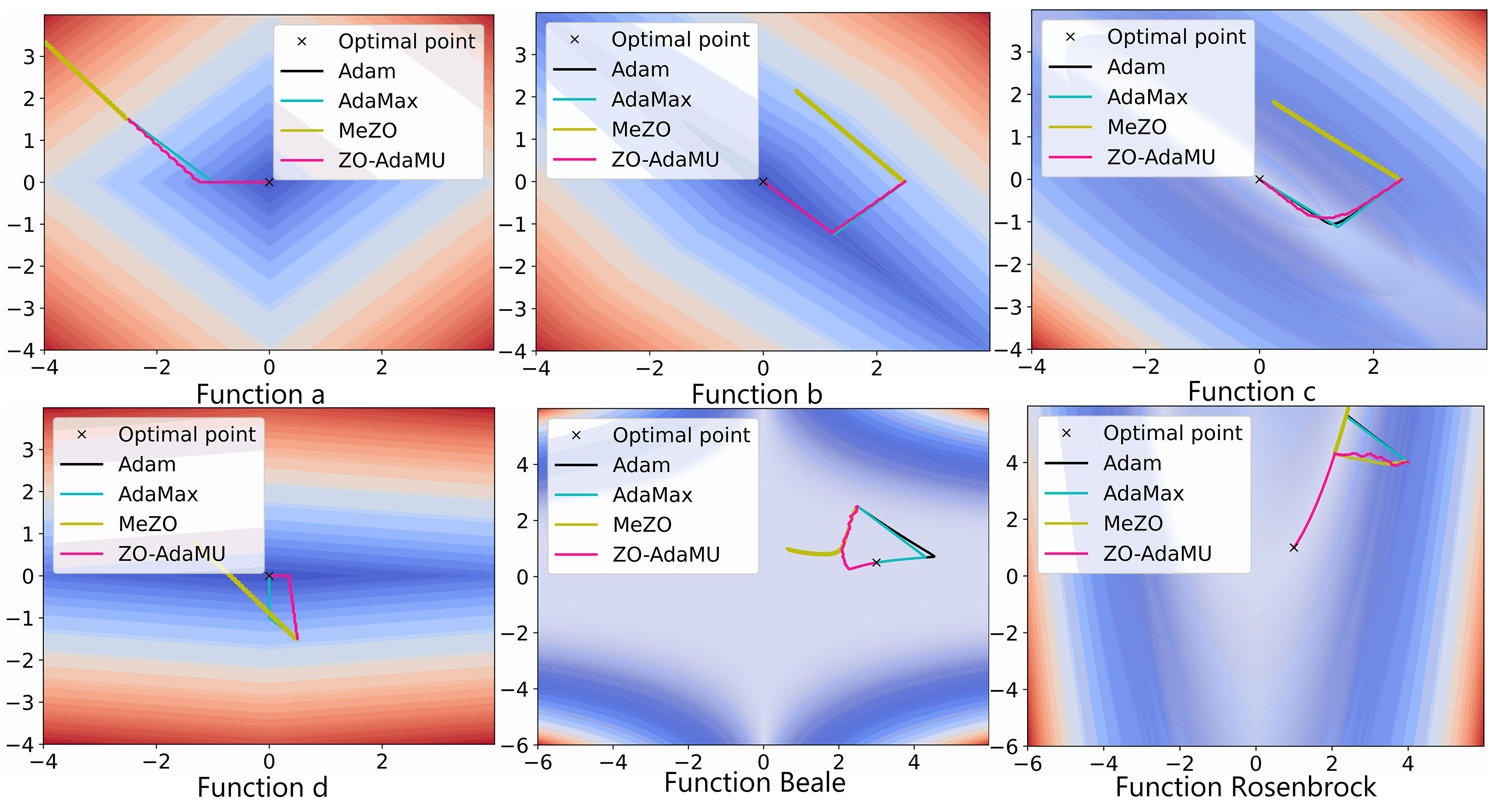}
    \caption{2D trajectories of Adam, AdaMax, MeZO and ZO-AdaMU on 6 test functions. In all cases, ZO-AdaMU performs comparably to that first-order optimization methods, like Adam and AdaMax, but MeZO does not reach optimal points.}
    \label{tab:trajectory}
\end{figure*}

\begin{figure}[htbp]
    \centering
    \includegraphics[width=7cm]{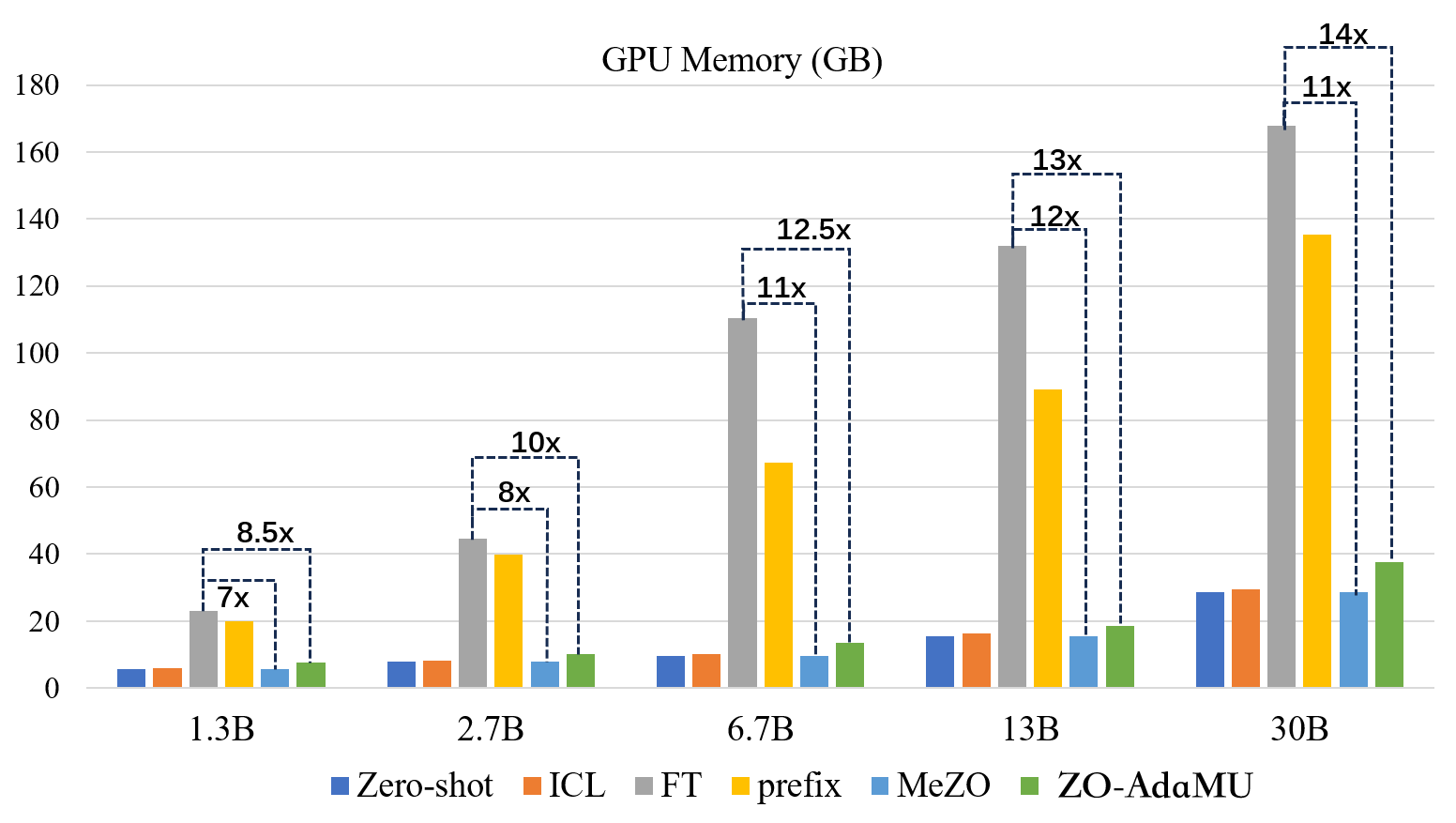}
    \caption{GPU memory consumption with OPT models fine-tuning on $2048$ maximum length per example.}
    \label{fig:memory}
\end{figure}

As the storage of the expected moving average (EMA) for perturbation momentum,
ZO-AdaMU slightly increases memory usage compared to MeZO.
In Figure \ref{fig:memory},
we summarize the memory usage of zero-shot, in-context learning (ICL), prefix learning and full-parameter fine-tuning with Adam, MeZO and ZO-AdaMU.
These statistics report the peak GPU memory usage by testing OPT models with Nvidia A100 GPUs on the SQuAD task (maximum length 2048 and minibatch size 1).

As shown in Figure \ref{fig:memory} that MeZO exhibits the same memory consumption as zero-shot,
which saves of up to 7 times of memory at least compared to back-propagation fine-tuning and 6 times compared to prefix fine-tuning.
Even though our proposed ZO-AdaMU has a slight of memory usage increase compared to MeZO,
it does not raise the requirements for mainstream hardware (like Nvidia A100 and V100) as shown in Table \ref{tab:memory}.
This advantage enables training larger models within a fixed hardware budget,
as illustrated in Figure \ref{fig:memory}.
Specifically,
using a single A100 GPU,
ZO-AdaMU allows for tuning a model that is 11 times larger than what is feasible with full-parameter fine-tuning.

\subsection{Trajectory Visualization on Test Functions}

In this section,
we validate the training trajectories of Adam, AdaMax, MeZO and ZO-AdaMU on 6 test functions,
and the 2D trajectories are shown in Figure \ref{tab:trajectory}.
These test functions are useful to evaluate characteristics of optimization algorithms, such as convergence rate, precision, robustness and general performance.

\begin{itemize}
    \item Function a: $f(x,y)=\left| x \right| + \left| y \right| $ with global minimum $f(0,0)=0$ and search domain $-3 \leqslant x,y \leqslant 3$;
    \item Function b: {\footnotesize $f(x,y)=\left| x+y \right| + \left| x-y \right|/10$} with global minimum $f(0,0)=0$ and search domain {\footnotesize $-3 \leqslant x,y \leqslant 3$};
    \item Function c: {\footnotesize $f(x,y)=(x+y)^2+(x-y)^2/10$} with global minimum $f(0,0)=0$ and search domain {\footnotesize $-3 \leqslant x,y \leqslant 3$};
    \item Function d: {\footnotesize $f(x,y)=\left| x \right|/10 + \left| y \right|$} with global minimum $f(0,0)=0$ and search domain {\footnotesize $-3 \leqslant x,y \leqslant$ 3};
    \item Function Beale : {\footnotesize $f(x,y)=(1.5-x+xy)^2+(2.25-x+xy^2)^2+(2.625-x+xy^3)^2$} with global minimum {\footnotesize $f(3,0.5)=0$} and search domain {\footnotesize $-4.5 \leqslant x,y \leqslant 4.5$};
    \item Function Rosenbrock: {\footnotesize $f(x,y)=100(x-y^2)^2+(1-y)^2$} with global minimum {\footnotesize $f(1,1)=0$} and search domain {\footnotesize $-4.5 \leqslant x,y \leqslant 4.5$}.
\end{itemize}

In all test functions,
adapting perturbation with momentum in ZO-SGD reaches optimal points on all test functions,
while MeZO optimizer fails to find any global minimum.
Compared with momentum-based back-propagation optimizers,
like Adam and AdaMax,
ZO-AdaMU shows similar trajectories and reaches the optimal points on test functions of (a), (b) and (c).
In addition,
on test functions of (d), Beale and Rosenbrock,
even though ZO-AdaMU shows different optimization trajectories,
it reaches the optimal points with a faster speed than Adam and AdaMax.

\section{Conclusion}

We propose a ZO-AdaMU optimizer in this work,
which adapts the momentum and uncertainty on simulated perturbation in the zeroth-order optimizer (ZO).
To our knowledge,
ZO-AdaMU is the first ZO-SGD optimizer that adapts the momentum on the stochastic approximation for simulated perturbation.
Even though storing perturbation momentum requires a little extra memory cost compared with MeZO,
ZO-AdaMU is consistent with MeZO for requirements of mainstream GPU hardware.
We experimentally validate that ZO-AdaMU outperforms MeZO and back-propagation optimizers on convergence rate and generalization across various NLP tasks.
Our visualizations prove that ZO-AdaMU performs comparably with Adam and AdaMax on popular test functions in machine learning.

\section{Acknowledgments}
This work is jointly supported by grants from the National Key R\&D Program of China (No. 2022ZD0116002),
the Project funded by China Postdoctoral Science Foundation (No. 2023M741843),
Shenzhen Science and Technology Plan (No. ShenKeJiChuangXinZhi[2023]87),
the Science and Technology Department of Guizhou Province (No. Qiankehe Support[2022]General019),
the National Social Science Foundation - Major Project (No. 20\&ZD226),
the Shenzhen Development and Reform Commission (No. XMHT20190108009),
the National Natural Science Foundation of China (No. 62276075, 62106115, 62006062 and 62176076),
the Guangdong Provincial Key Laboratory (No. 2022B1212010005),
the Major Key Project of PCL (No. PCL2022D01, PCL2023A09),
the Key Laboratory of Intelligent Computing in Network Environment.

\bibliography{aaai24}

\end{document}